\documentclass[letterpaper, 10 pt, conference]{ieeeconf}  

\usepackage{microtype}
\usepackage{graphicx}
\usepackage{subcaption}
\usepackage{booktabs} 
\usepackage{amsmath}
\makeatletter\let\labelindent\undefined\makeatother
\usepackage{enumitem}
\usepackage{cite}
\usepackage{xurl}
\usepackage{amsfonts}

\IEEEoverridecommandlockouts                              

\title{\LARGE \bf
PLS-Calib: A Partial Least Squares Framework for Event Camera and Odometry Calibration under Ground Motion Constraints
}

\author{Guangyu Li$^{1}$, Xiao Li$^{2}$, Yujie Wu$^{3}$, Changshuo Wang$^{4}$, \\ Prayag Tiwari$^{5}$, Jiang Cai$^{1}$, Fangwen Yu$^{6,\dagger}$, and Mingkun Xu$^{1,\dagger}$
\thanks{$^{\dagger}$Corresponding authors: Fangwen Yu and Mingkun Xu.}%
\thanks{This work was supported in part by the High-Level Talents Innovation Team Project of the Guangdong-Macao In-Depth Cooperation Zone in Hengqin under Grant No.~2630004018939, in part by the National Natural Science Foundation of China (NSFC) under Grant No.~62506084, in part by the Natural Science Foundation of Jiangsu Province under Grant No.~BK20243064, and in part by the Beijing Natural Science Foundation under Grant No.~4262048.}%
\thanks{$^{1}$Guangyu Li, Jiang Cai, and Mingkun Xu are with the Guangdong Institute of Intelligence Science and Technology, Zhuhai, China. {\tt\small xumingkun@gdiist.cn}}%
\thanks{$^{2}$Xiao Li is with the State Key Laboratory of Deep-Sea Science and Intelligent Technology,Institute of Deep-sea Science and Engineering, Chinese Academy of Sciences.}%
\thanks{$^{3}$Yujie Wu is with The Hong Kong Polytechnic University, Hong Kong SAR, China.}%
\thanks{$^{4}$Changshuo Wang is with University College London, London, United Kingdom.}%
\thanks{$^{5}$Prayag Tiwari is with Halmstad University, Halmstad, Sweden.}%
\thanks{$^{6}$Fangwen Yu is with Tsinghua University, Beijing, China. {\tt\small yufangwen@tsinghua.edu.cn}}%
}

\makeatletter
\newcommand\blfootnote[1]{%
  \begingroup
  \renewcommand\thefootnote{}\footnote{#1}%
  \addtocounter{footnote}{-1}%
  \endgroup
}
\makeatother

\begin{document}

\maketitle
\thispagestyle{empty}

\blfootnote{Accepted at the 2026 IEEE/RSJ International Conference on Intelligent Robots and Systems (IROS 2026), Pittsburgh, PA, USA.}
\blfootnote{\copyright~2026 IEEE. Personal use of this material is permitted. Permission from IEEE must be obtained for all other uses, in any current or future media, including reprinting/republishing this material for advertising or promotional purposes, creating new collective works, for resale or redistribution to servers or lists, or reuse of any copyrighted component of this work in other works.}
\pagestyle{empty}

\begin{abstract}

Accurate extrinsic rotation calibration between sensors is fundamental to the performance of robotic perception systems. However, most existing calibration techniques rely on full 6-DoF motion to excite all degrees of freedom, which is often infeasible for ground-constrained robots with limited motion capabilities. Recent approaches designed for such restricted settings, such as Canonical Correlation Analysis (CCA)-based methods, suffer from ill-conditioned covariance matrices that lead to numerical instability and suboptimal calibration accuracy.
To overcome these limitations, we present a novel rotation calibration framework named PLS-Calib that, for the first time, leverages Partial Least Squares (PLS) regression to model the latent kinematic correlations between asynchronous, heterogeneous sensor streams. Specifically, we apply our method to the calibration of an event camera and an odometry onboard a ground robot. To improve event-based pattern detection, we introduce a polarity-aware event representation, which enhances spatiotemporal contrast in circular calibration targets. Our PLS-based formulation yields a closed-form, stable solution that avoids matrix singularities inherent in CCA-based approaches.
Extensive experiments on both synthetic and real-world datasets validate the effectiveness of our approach, demonstrating significant improvements in calibration robustness and accuracy over state-of-the-art methods. This work offers a practical and theoretically grounded solution for rotation calibration in constrained robotic systems and opens up new directions for applying statistical learning techniques in neuromorphic vision.

\end{abstract}

\section{INTRODUCTION}

A key prerequisite for effective multi-sensor fusion is rotation calibration, which estimates the spatial transformation between different sensing modalities. In systems that combine motion and vision sensors, such as event cameras and odometry, precise calibration is essential to ensure consistent spatial alignment, while misalignment can lead to degraded perception accuracy and unstable control behavior \cite{zhang2004extrinsic, beltran2022automatic}.

Event cameras represent a paradigm shift in visual sensing. Inspired by the biological retina, they asynchronously capture per-pixel brightness changes and generate a sparse stream of events encoding location, time, and polarity \cite{gallego2020event}. Compared to conventional frame-based cameras, event cameras offer ultra-low latency, low redundancy, and high dynamic range (HDR), making them particularly suitable for real-time robotic perception in dynamic and challenging environments. However, their fundamentally different data structure poses new challenges for computer vision and robotic calibration, where conventional feature-based pipelines are often inapplicable. In particular, classical calibration procedures relying on static checkerboard detection become unreliable, since event cameras only respond to brightness changes and do not directly capture absolute intensity patterns. As a result, stable corner extraction requires controlled motion excitation, and feature detection is highly sensitive to motion speed, contrast, and noise \cite{muglikar2021calibrate, huang2021dynamic}.

\begin{figure}[t]
    \centering
    \includegraphics[width=0.55\columnwidth]{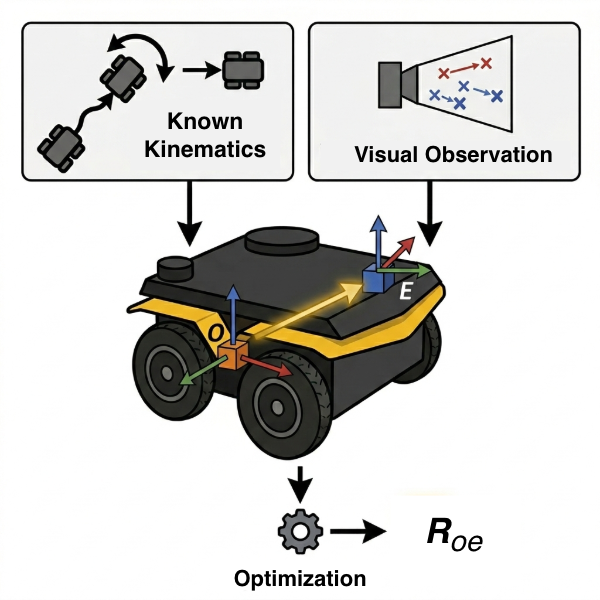}
    \caption{Overview of the proposed extrinsic calibration configuration.}
    \label{fig:calib_framework}
\end{figure}

Extrinsic calibration between event cameras and odometry sensors is especially non-trivial in ground-constrained settings, where robots move primarily on planar surfaces. Most prior calibration methods rely on full 6-DoF motion to ensure observability of the transformation, which is often infeasible for such planar robots. Moreover, CCA, which is commonly used for trajectory alignment based calibration, can suffer from numerical instability when the motion lacks sufficient diversity, leading to singular covariance matrices, which result in unreliable calibration or the need for external calibration rigs \cite{yin2021m2dgr,li2024spatio}. While recent learning-based methods offer alternatives, they demand heavy supervision, computational resources, and may lack generalizability to unseen setups \cite{liao2023deep}.

To overcome these limitations, we propose PLS-Calib, a novel rotation calibration framework based on PLS regression. Unlike CCA, PLS naturally handles rank-deficient data by projecting observations into a shared latent space, maximizing covariance rather than correlation. This enables robust, closed-form estimation of the transformation between event camera and odometry frames, even under restricted planar motion. Our method supports calibration under ground-constrained (planar) motion, making it ideal for practical deployment in ground-based robotic systems. We further introduce a polarity-aware event representation to enhance spatiotemporal encoding, enabling more stable pose estimation in the event-based vision pipeline \cite{wold1984collinearity,de1993simpls,macgregor1994process,blanco2000nir,spiegelman1998theoretical,rosipal2005overview}. Experiments on both synthetic and real-world datasets demonstrate that PLS-Calib outperforms CCA-based and neural calibration baselines in accuracy and robustness, especially under motion constraints. To the best of our knowledge, our work is the first to apply PLS in event camera calibration, replacing the ill-conditioned whitening step of CCA-based methods and providing a computationally efficient, biologically motivated solution for sensor fusion in modern robotics.


Overall, our method capitalizes on the temporal and spatial characteristics of event-based vision to perform precise calibration under ground-constrained motion (Fig.~\ref{fig:calib_framework}). Specifically, we employ an event camera to detect circular calibration patterns and estimate time-synchronized robot poses. These are then paired with corresponding odometry data to construct a PLS regression model that estimates the optimal extrinsic rotation matrix while explicitly respecting kinematic constraints. This formulation mitigates the common issue of matrix singularities encountered in planar motion calibration scenarios.

The key contributions of this work are summarized as follows:

\begin{itemize}[leftmargin=*, topsep=0pt, itemsep=0pt, parsep=0pt]
\item We introduce \textbf{PLS-Calib}, a novel kinematics-aware spatio-temporal calibration framework that leverages PLS regression to robustly estimate extrinsic rotation parameters from event camera and odometry data, addressing degeneracies arising from restricted ground-plane motion.
\item We propose a polarity-enhanced Time Surface representation that improves feature detection stability in event data, facilitating accurate localization of circular calibration patterns under high-speed, low-light, or HDR conditions.
\item We extend pattern-based vision calibration to the domain of event-based sensor fusion, achieving accurate and target-based rotation calibration without requiring full 6-DoF excitation. Our approach demonstrates strong performance in both synthetic and real-world robotic platforms, offering a lightweight, generalizable solution for sensor alignment in constrained environments.
\end{itemize}

\section{RELATED WORK}

In a robot perception system with multiple fused sensors, extrinsic calibration of the sensors is required for the robot operation to accurately correspond to the input data of the sensors and to map the measured values to the coordinate space in which the robot is working; the main methods are the hand-eye calibration method, the learning-based calibration method, and the kinematic correlation-based method. Among them, the hand-eye calibration method is the early classical method, hand-eye calibration by solving the coordinate transformation relationship between two sensors: AX=XB, where A and B are the change matrices between the two sensor poses, respectively, and X is the pose transformation relation matrix to be solved. The optimal extrinsic calibration of the sensors is solved as the transformation relation X by corresponding the characteristics of the measured data to the sensor data.\cite{tsai1989new,strobl2006optimal,horaud1995hand}

Traditional hand-eye calibration methods rely on precision geometric calculations and algebraic solutions, which in practical applications require input sensors to maintain a high degree of spatio-temporal synchronisation and spatial consistency, and will lead to a decrease in calibration accuracy if noise or measurement errors exist in complex environments. With the development of deep learning, the above problems can also be solved by deep learning. For example, by applying convolutional neural network (CNN) and graph neural network (GNN)\cite{liao2023deep}, an end-to-end calibration network model can be realised, which is able to maintain a certain degree of robustness in complex scenes and dynamic changes and can be a solution for the calibration problems of camera internal reference, external reference and aberration, and even the data calibration problems of multiple sensor fusion systems. For bio-inspired event cameras, using a spiking neural network (SNN) to process the time-series of event streams is also a good way to solve the problem, and it has been well applied in both velocity estimation problems\cite{yang2016monocular,li2025neurove,tian2023egomotion,orchard2013spiking}.

However, learning-based methods for sensor calibration require good and applicable datasets and sensor schemes. The motion correlation-based method for calibrating external sensor parameters has low data requirements and high computational efficiency, and can be used for a wide range of sensors to achieve sensor calibration. Qiu et al.~\cite{qiu2020real} proposed a unified calibration framework based on 3D motion correlation that efficiently computes the temporal offset and extrinsic rotation between two heterogeneous sensors. Building on this line, Li et al.~\cite{li2024spatio} presented a spatio-temporal calibration method for an omnidirectional-vehicle event camera, solving the ground-robot calibration with a CCA scheme. However, most of these motion-correlation methods rely on CCA, whose covariance matrix becomes singular under ground-constrained motion, biasing the estimated extrinsics.

Usually, to avoid the result bias brought by the ground constraint, the researcher needs to fix the sensors to the holder for calibration, but this method may have the possibility of some error during the calibration process. Li et al.\ estimate the temporal offset and then obtain the extrinsic rotation directly from the covariance matrices produced by the CCA process; however, this CCA-based formulation still relies on covariance whitening, which becomes ill-conditioned under planar motion. Kim et al. \cite{kim2024gril} proposed the GRIL-Calib method to fix the odometry and LiDAR sensors fixed on a moving trolley. GPM constraints are proposed and incorporated into the optimisation process to avoid the problem of the ground trolley not being able to acquire the complete 6-degree-of-freedom extrinsic calibration due to motion constraints.

\section{METHODOLOGY}

\begin{figure}[t]
    \centering
    
    \begin{subfigure}{\columnwidth}
        \centering
        \includegraphics[width=\linewidth]{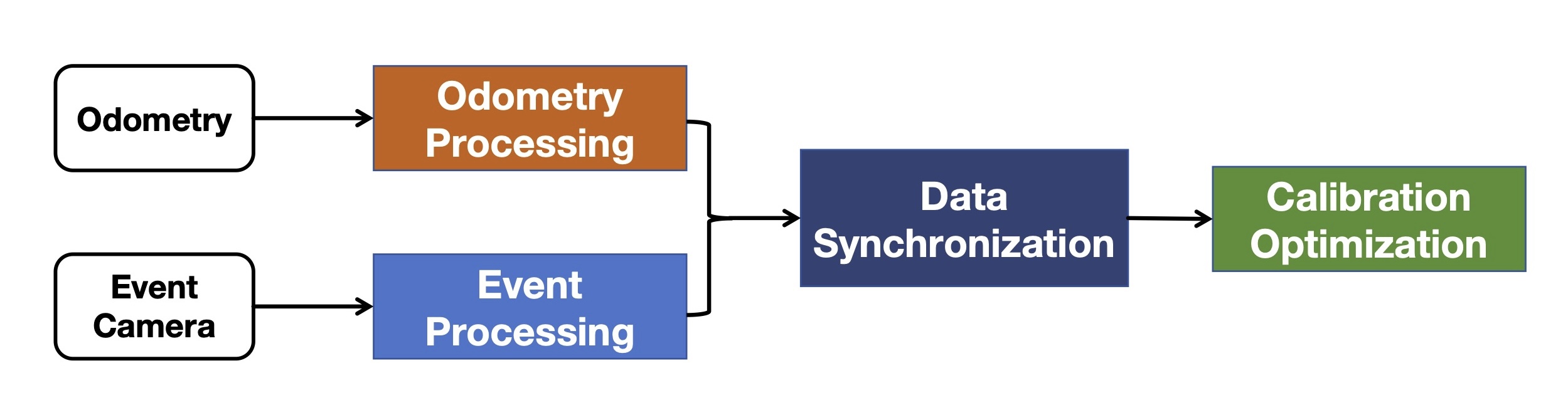}
        \caption{Overview of the proposed PLS-Calib framework.}
        \label{fig:framework}
    \end{subfigure}
    
    \vspace{3pt}
    
    \begin{subfigure}{0.48\columnwidth}
        \centering
        \includegraphics[width=\linewidth]{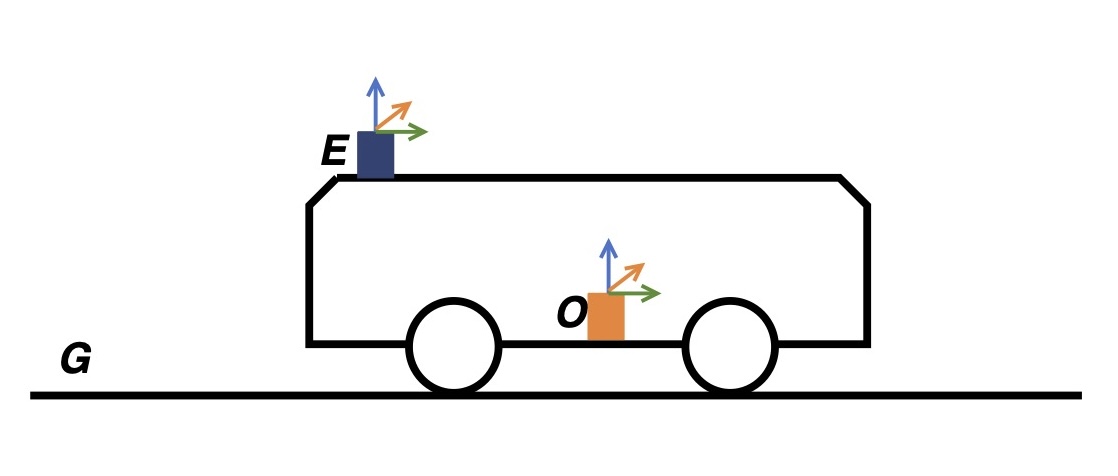}
        \caption{PLS-Calib robot platform.}
        \label{fig:platform}
    \end{subfigure}
    \hfill
    \begin{subfigure}{0.44\columnwidth}
        \centering
        \includegraphics[width=\linewidth]{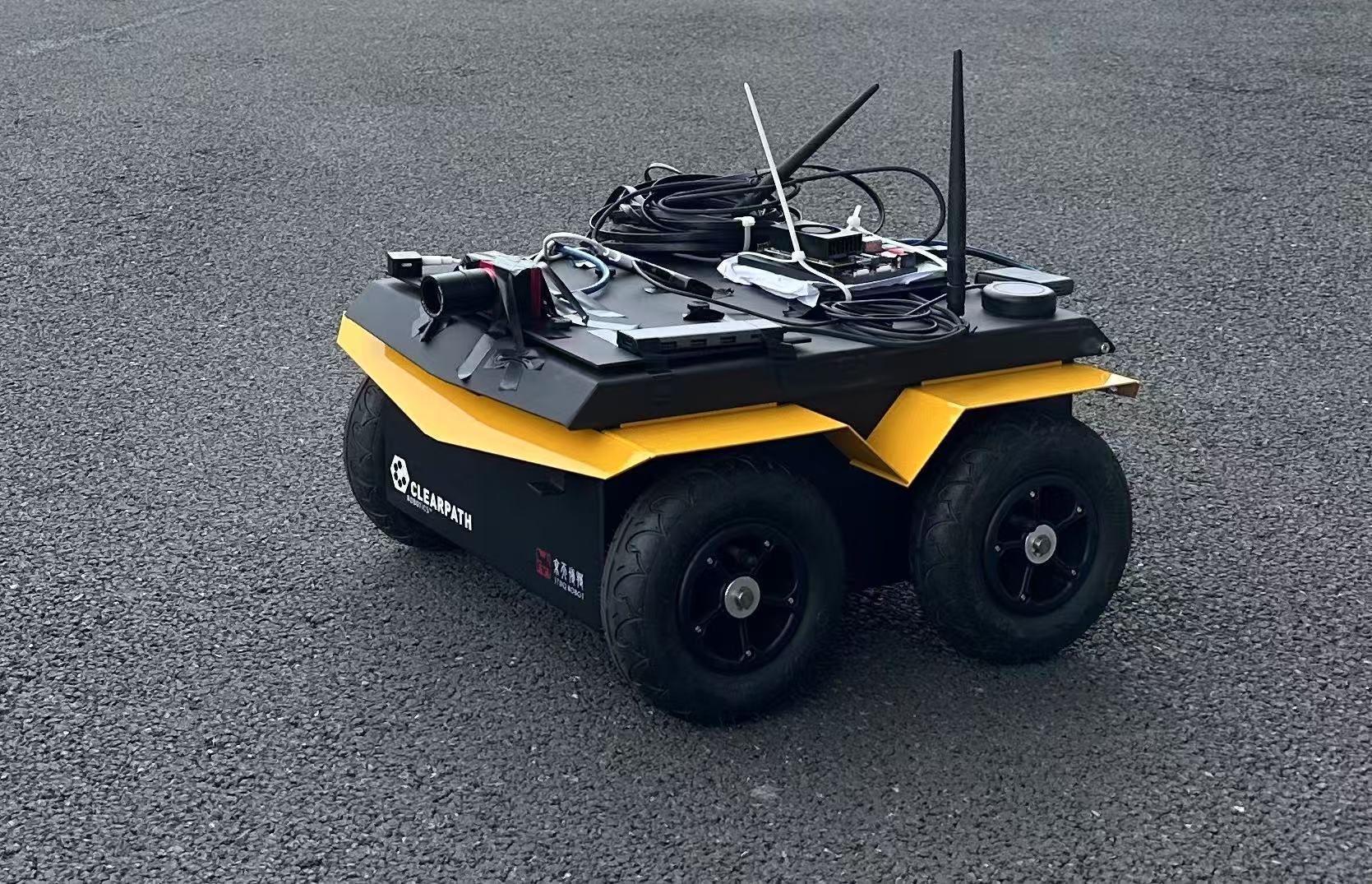}
        \caption{Experimental robot setup.}
        \label{fig:robot}
    \end{subfigure}

    \caption{Overview of the proposed PLS-Calib system and experimental platform. (a) Calibration framework. (b) Robot platform. (c) Experimental setup.}
    \label{fig:pls_system}
\end{figure}

We propose PLS-Calib, a calibration framework specifically designed for ground-constrained robots as shown in Fig.~\ref{fig:pls_system}. In our platform, a ground robot is equipped with an event camera and an odometry sensor to simultaneously capture trajectory data during motion. The primary objective of this framework is to estimate the optimal transformation between the two heterogeneous sensor modalities and recover their temporal offset. PLS-Calib addresses the challenge of calibrating robots under ground motion constraints, where full 6-degree-of-freedom movement is not possible. It effectively reduces the bias commonly observed in existing CCA-based calibration methods.

\subsection{Trajectory from Event Data}

Event cameras are bio-inspired visual sensors that differ from conventional RGB frame cameras in their imaging principle: each pixel operates independently and asynchronously triggers an event when the change in light intensity exceeds a threshold. Each event $e_k = (x_k, y_k, p_k, t_k)$ encodes the pixel location, timestamp, and polarity, and is generated only when there is relative motion between the camera and the scene.

Due to their asynchronous data generation mechanism, the visual information representation of event cameras significantly differs from that of standard RGB cameras \cite{zhou2021event}. A common method to visualise event data is to accumulate events over a short time window and sum their polarities into a basic event frame. However, this approach discards the temporal resolution inherent in event streams. To address this limitation, the Time Surface (TS) representation was introduced, which records the timestamp of the most recent event at each pixel to preserve fine-grained motion dynamics.

Time surface \cite{nagata2021optical, manderscheid2019speed, lagorce2016hots} is a common event data representation method to convert events into an image. The core of the Time Surface $S_i$ is to map each event into a 2d-pixel position. Additionally,  using a two-dimensional array to represent each event pixel $T_i(u, p)$, where the value of each pixel denotes the timestamp $t$ difference since the last event occurred with polarity $p$ at that pixel,  can effectively preserve both the temporal and spatial information of the events.

\begin{equation}
   S_i(u, p) = e^{ -\frac{ t_i - T_i(u,p) }{ \tau } }.
\end{equation}

Here $\tau$ is the time constant for the values of $T_i(u,p)$ in this exponential decay kernel.

\begin{figure}[t]
    \centering

    \begin{subfigure}{0.32\columnwidth}
        \centering
        \includegraphics[width=\linewidth]{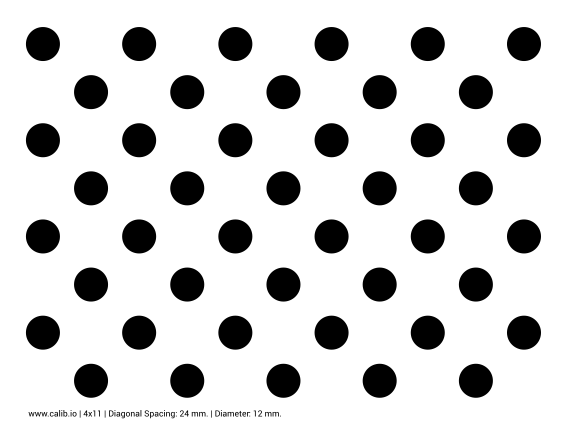}
        \caption{}
        \label{fig3a}
    \end{subfigure}
    \hfill
    \begin{subfigure}{0.32\columnwidth}
        \centering
        \includegraphics[width=\linewidth]{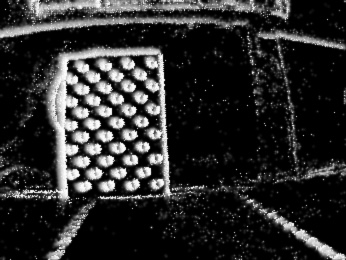}
        \caption{}
        \label{fig3b}
    \end{subfigure}
    \hfill
    \begin{subfigure}{0.32\columnwidth}
        \centering
        \includegraphics[width=\linewidth]{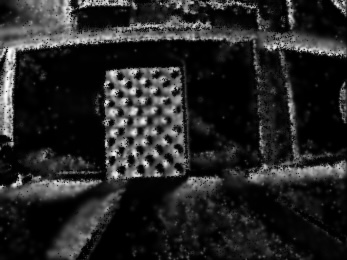}
        \caption{}
        \label{fig3c}
    \end{subfigure}

    \caption{Representation of event data. 
    (a) Asymmetric circle grid. 
    (b) Time Surface representation. 
    (c) Proposed representation.}
    \label{fig3}
\end{figure}

With the representation of event data, we apply the calibration pattern to estimate the pose of the robot. For the event stream data acquired by the event camera, we need to represent the acquired event data as image frames in order to acquire the corresponding features for motion trajectory estimation. Due to the imaging characteristics of the event camera, the conventional checkerboard grid may lose a small amount of data in calibration when the trajectories are parallel to the edges of the grid. Based on previous studies, we use a circular calibration pattern (Fig.~\ref{fig3a}) for imaging and pose estimation from the event camera, using Perspective-n-Point (PnP) to solve the pose estimation and obtain the motion trajectory.

However, according to the Time Surface method above, we obtain the circular calibration pattern in event representation (Fig.~\ref{fig3b}). However, the time-related decay mechanism in the Time Surface approach can lead to trailing artifacts in motion imagery. Therefore, we improved the decay mechanism of the Time Surface. Instead of updating each pixel based on the time difference, we modified the method to update based on the difference in event polarity \( p \). After each update, we apply a Gaussian blur to the Time Surface to smooth it and reduce noise. This modification reduced the presence of event data trailing artifacts, as demonstrated in Fig.~\ref{fig3c}, and enabled more accurate feature recognition of the calibration board in computer vision methods. 
We have,
\begin{equation}
   S_i(u, p) = e^{ -\frac{ \left| p_i - p_{i-1} \right| }{ \alpha } },
\end{equation}
where $\alpha=0.1$ is a dimensionless scaling constant controlling the sharpness of the polarity-based decay; unlike the time constant $\tau$ in Eq.~(1), it does not carry units of seconds. A $5\times5$ Gaussian blur ($\sigma=0.7$) is then applied to smooth the surface.

 In the previous Time Surface approach, motion and occlusion often caused the circular markers on the calibration plate to appear distorted or blurred. Leveraging the polarity-based event update mechanism, our method effectively reduces motion-induced blurring in the representation of event data. This improvement is critical for accurately detecting the circular features of the calibration pattern. Our approach mitigates these issues, even in the presence of temporal decay effects inherent to event-based sensing, thereby significantly enhancing the robustness of circle detection during robot movement.

\subsection{Trajectory from Odometry}

To obtain the velocity of motions, we leverage the wheel odometry available on the Jackal Clearpath robot for accurate motion tracking. In a differential-drive mobile robot, wheel odometry estimates the robot’s planar motion by computing its linear and angular velocities from the rotational displacements of the left and right wheels.

\begin{equation}
\begin{aligned}
v_L &= r \cdot \omega_L, \\
v_R &= r \cdot \omega_R. \\
\end{aligned}
\end{equation}

Given the linear velocities of the left and right wheels, the robot’s translational and angular velocities can be computed as:
\begin{equation}
\begin{aligned}
v &= \frac{v_R + v_L}{2}, \
\omega &= \frac{v_R - v_L}{b},
\end{aligned}
\end{equation}
where $v_R$ and $v_L$ denote the velocities of the right and left wheels, respectively, and b is the distance between the two wheels.

Using these equations above, we derive the robot’s motion trajectory from the odometry data. This trajectory is later aligned with the motion trajectory estimated by the event camera in order to compute the rotation matrix between the two sensor frames during movement.

There are numerous methods for robots to achieve motion velocity estimation, our method can also be generalized to other sensors capable of detecting motion velocity. In general, IMUs are widely used for estimating motion, velocity, and heading direction in mobile robotics applications~\cite{yang2016monocular, feng2024s3e,chen2023esvio, helmberger2022hilti}. The gyroscope and accelerometer measurements provided by the IMU are used to compute angular velocity and linear acceleration, which assist in inferring the robot's orientation and motion.

\begin{equation}
\begin{aligned}
\hat{\omega}_i &= \omega_i + b_g + \varepsilon_g, 
\\
\hat{a}_i &= a_i + R_{I_i}^w \cdot g_w + b_a + \varepsilon_a.
\end{aligned}
\end{equation}


However, IMU measurements are prone to drift over time, which can lead to significant errors in long-term trajectory estimation. For this reason, we did not adopt an IMU-based estimation approach in our experimental setup.

\subsection{Extrinsic Rotation}

The goal of extrinsic rotation is to estimate the extrinsic rotation matrix $R_{oe}$ between the two sensors. Given the velocity measurements from the event camera and odometry, denoted as $V_e$ and $V_o$, respectively, we aim to determine the rotational relationship that aligns the two motion representations.

\begin{equation}
    V_o = R_{oe} V_e.
\end{equation}

Comparison of PLS regression methods for CCA In traditional sensor calibration work, the CCA method is mainly used to solve for the maximum correlation between the velocities measured by the two sensor data, so as to obtain the optimal sensor transformation relationship. However, in robots subject to ground constraints, solving with the CCA method can lead to deviations in the final results due to matrix singularities, yielding unstable calibration results. Therefore, we use the method of constructing a PLS regression equation to determine the time offset and extrinsic rotation between the two sensors to circumvent the problem of matrix singularity.

\subsubsection{Calibration via Canonical Correlation Analysis}

The Canonical Correlation Analysis (CCA) method is widely used in prior calibration studies. It is a statistical approach for evaluating the correlation between two data sets, such as sensor data streams $X$ and $Y$, by finding their respective linear combinations $a^T X$ and $b^T Y$ that are maximally correlated. The correlation coefficient, commonly used as the metric, is defined as $\rho = \frac{\text{Cov}(X, Y)}{\sqrt{\text{Var}(X)} \sqrt{\text{Var}(Y)}}$, where $\text{Cov}(X, Y)$ denotes the covariance between $X$ and $Y$, and $\text{Var}(X)$ and $\text{Var}(Y)$ represent their variances.

In the actual solution process, the event data and odometry data we acquired need to be unified into one-dimensional data for the calculation of correlation coefficients. We have $U = Xa$ and $V = Yb$.
The optimisation objective of the CCA method is to maximise the correlation coefficient, which is,

\begin{equation}
\arg\max_{a,b} \frac{\text{Cov}(U, V)}{\sqrt{\text{Var}(U) \text{Var}(V)}}.
\end{equation}

Let \( Z = \begin{bmatrix} X \\ Y \end{bmatrix} \). Then we have the expectation
\( E(Z) = \begin{bmatrix} E(X) \\ E(Y) \end{bmatrix} \),
and the covariance matrix
\( \Sigma = \text{Var}(Z) = \begin{bmatrix} \Sigma_{11} & \Sigma_{12} \\ \Sigma_{21} & \Sigma_{22} \end{bmatrix} \).
The variances and covariance of the linear projections are given by
\( \text{Var}(U) = a^T \text{Cov}(X) a = a^T \Sigma_{11} a \),
\( \text{Var}(V) = b^T \text{Cov}(Y) b = b^T \Sigma_{22} b \), and
\( \text{Cov}(U, V) = a^T \text{Cov}(X, Y) b = a^T \Sigma_{12} b \).

When subjected to ground constraints, the $z$-axis of our acquired data becomes zero. Let
\[
X = [x_1, x_2, 0]_{n \times 3}, \quad
Y = [y_1, y_2, y_3]_{n \times 3},
\]
we define the combined data matrix as
\[
Z = [x_1, x_2, 0, y_1, y_2, y_3]_{n \times 6}.
\]
Therefore, the variance of $Z$ is computed as
\begin{equation}
\text{Var}(Z) = E[(Z - \mu_z)^T (Z - \mu_z)],
\end{equation}
where $\mu_z$ denotes the mean of $Z$.
Let 
\(\hat{Z} = [x_1, x_2, 0, y_1, y_2, y_3]_{n \times 6} = Z - \mu_z\),
We have,
\begin{equation}
\begin{aligned}
\text{Var}(Z) &= \frac{1}{n} \hat{Z}^T \hat{Z} \\
&= \frac{1}{n} [x_1, x_2, 0, y_1, y_2, y_3]^T [x_1, x_2, 0, y_1, y_2, y_3] \\
&= \frac{1}{n}
\begin{bmatrix}
x_1^T x_1 & x_1^T x_2 & 0 & x_1^T y_1 & x_1^T y_2 & x_1^T y_3 \\
x_2^T x_1 & x_2^T x_2 & 0 & x_2^T y_1 & x_2^T y_2 & x_2^T y_3 \\
0 & 0 & 0 & 0 & 0 & 0 \\
y_1^T x_1 & y_1^T x_2 & 0 & y_1^T y_1 & y_1^T y_2 & y_1^T y_3 \\
y_2^T x_1 & y_2^T x_2 & 0 & y_2^T y_1 & y_2^T y_2 & y_2^T y_3 \\
y_3^T x_1 & y_3^T x_2 & 0 & y_3^T y_1 & y_3^T y_2 & y_3^T y_3
\end{bmatrix} \\
&=
\begin{bmatrix}
\Sigma_{11} & \Sigma_{12} \\
\Sigma_{21} & \Sigma_{22}
\end{bmatrix}
\end{aligned}.
\end{equation}


Since the third channel of $X$ is constant, $S_{xx}$ becomes rank-deficient (i.e., $\mathrm{rank}(S_{xx}) \le 2$), making the whitening step $S_{xx}^{-1/2}$ ill-conditioned and leading to unstable canonical directions.

\subsubsection{Calibration via Partial Least Squares Regression}

To address the limitations of CCA under ground motion constraints, we adopt a Partial Least Squares (PLS) regression strategy to calibrate the event camera and odometry sensor based on their observed velocity data. Unlike CCA, which seeks maximally correlated projections and requires covariance whitening (potentially ill-conditioned under planar motion), PLS constructs latent components by maximizing the \emph{covariance} between projected variables and then performs regression in the latent space.

Let $X,Y\in\mathbb{R}^{n\times 3}$ denote the synchronized velocity measurements from the event camera and odometry, respectively, where each row stores one velocity sample as a row vector. We model their relationship as
\begin{equation}
Y = XR + E,
\label{eq:pls_model}
\end{equation}
where $R\in\mathbb{R}^{3\times 3}$ is the regression coefficient matrix and $E$ is the residual. Under this row-vector convention, the per-sample relation in Eq.~(6) reads $Y = X R_{oe}^{\top} + E$, i.e., the regression coefficient matrix corresponds to the transpose of the extrinsic rotation, $R = R_{oe}^{\top}$.

\paragraph{Standardization.}
Following the implementation, we standardize each channel of $X$ and $Y$ (column-wise) as
\begin{equation}
\tilde{X} = \frac{X-\mu_X}{\sigma_X+\varepsilon},\quad \tilde{Y} = \frac{Y-\mu_Y}{\sigma_Y+\varepsilon},
\end{equation}
where $\mu_X,\mu_Y\in\mathbb{R}^{1\times 3}$ are the column means, $\sigma_X,\sigma_Y\in\mathbb{R}^{1\times 3}$ are the column standard deviations, and $\varepsilon$ is a small positive constant ($10^{-10}$ in our implementation) that prevents division by zero for degenerate channels, e.g., the $z$-axis velocity that remains constant under planar motion.

\paragraph{Latent direction estimation.}
PLS seeks a direction $w\in\mathbb{R}^{3}$ such that the score $t=\tilde{X}w$ has maximum covariance with the multivariate response $\tilde{Y}$. In our implementation, this is achieved by maximizing the squared covariance magnitude
\begin{equation}
\begin{aligned}
w &= \arg\max_{\|w\|=1}\ \big\|\tilde{Y}^T(\tilde{X}w)\big\|_2^2 \\
  &= \arg\max_{\|w\|=1}\ w^T\big(\tilde{X}^T\tilde{Y}\tilde{Y}^T\tilde{X}\big)w,
\end{aligned}
\label{eq:pls_cov_obj}
\end{equation}
which leads to the eigenvalue problem
\begin{equation}
\big(\tilde{X}^T\tilde{Y}\tilde{Y}^T\tilde{X}\big)\,w = \lambda w.
\label{eq:pls_eig}
\end{equation}
The eigenvector corresponding to the largest eigenvalue is selected as the projection direction. The latent score is then computed as
\begin{equation}
t=\tilde{X}w.
\label{eq:pls_score}
\end{equation}

\paragraph{Deflation and multiple components.}
To extract multiple components, we iteratively remove from the standardized input the part explained by the current score. Let $E_0=\tilde{X}$. For the current score $t$, we compute the loading
\begin{equation}
p = \frac{E^T t}{t^T t},
\label{eq:pls_loading}
\end{equation}
and update
\begin{equation}
E \leftarrow E - tp^T.
\label{eq:pls_deflation}
\end{equation}
Repeating the procedure yields $K$ components. Stacking the scores gives $T=[t_1,\dots,t_K]\in\mathbb{R}^{n\times K}$.

\paragraph{Regression in the latent space.}
Given the score matrix $T$, we estimate the regression from scores to the standardized response by least squares,
\begin{equation}
\tilde{Y} \approx TC,\qquad C = T^\dagger \tilde{Y}.
\label{eq:pls_score_reg}
\end{equation}
To map the regression back to the standardized input space, we use a corrected weight matrix $W^\star\in\mathbb{R}^{3\times K}$, which is equivalent to $W^\star = W(P^T W)^{-1}$ with $W=[w_1,\dots,w_K]$ and $P=[p_1,\dots,p_K]$, and obtain
\begin{equation}
R_{\mathrm{std}} = W^\star C.
\label{eq:pls_Rstd}
\end{equation}
The regression matrix $R$ in the original (unstandardized) scale is then recovered from $R_{\mathrm{std}}$ by undoing the column-wise standardization in Eq.~(11).

\paragraph{Rotation construction.}
The regression matrix $R$ is not guaranteed to be orthogonal. In our implementation, instead of projecting $R$ to the nearest element in $SO(3)$ via SVD, we construct a right-handed frame from the first two row vectors of $R$.
Let $r_1^T = R_{1,:}$ and $r_2^T = R_{2,:}$. We compute
\begin{equation}
\hat{r}_1=\frac{r_1}{\|r_1\|},\qquad
\hat{r}_2=\frac{r_2}{\|r_2\|},\qquad
\hat{r}_3=\frac{\hat{r}_1\times\hat{r}_2}{\|\hat{r}_1\times\hat{r}_2\|},
\label{eq:pls_cross}
\end{equation}
and assemble
\begin{equation}
R_{\mathrm{rot}} = \big[\ \hat{r}_1\ \hat{r}_2\ \hat{r}_3\ \big].
\label{eq:pls_Rrot}
\end{equation}

Since the regression matrix satisfies $R = R_{oe}^{\top}$, its rows correspond to the columns of $R_{oe}$; normalizing the first two rows and completing the right-handed frame via the cross product in Eq.~(20) therefore directly recovers the extrinsic rotation, i.e., $R_{\mathrm{rot}}$ is the final estimate of $R_{oe}$ in the convention of Eq.~(6).


As discussed above, while CCA identifies maximally correlated projections, it can become numerically unstable under ground-constrained motion due to rank-deficient covariance matrices and ill-conditioned whitening. In contrast, the above PLS procedure constructs latent components by covariance maximization and performs regression via pseudoinverses in the latent space, making it more robust in planar-motion scenarios. This is particularly suitable for aligning the velocity measurements of an event camera and an odometry sensor for extrinsic rotation calibration.

\section{EXPERIMENTS AND RESULTS}

The experiment is mainly composed of two parts. Firstly, we validate the calibration results of the PLS method under ground constraints in the synthesised data and compare the calibration results of the CCA method. Secondly, we built a physical ground robot to collect the data and complete the calibration validation in the real world using a robot with odometry and an event camera.

\subsection{Synthetic Data Generation}

In order to evaluate the effectiveness of our method under ground constraints, we generated synthetic datasets for two scenarios: one with full 6-DoF motion, and the other constrained to planar motion, simulating typical ground-based robot behavior.

\begin{figure}[t]
    \centering

    \begin{subfigure}{0.46\columnwidth}
        \centering
        \includegraphics[width=\linewidth]{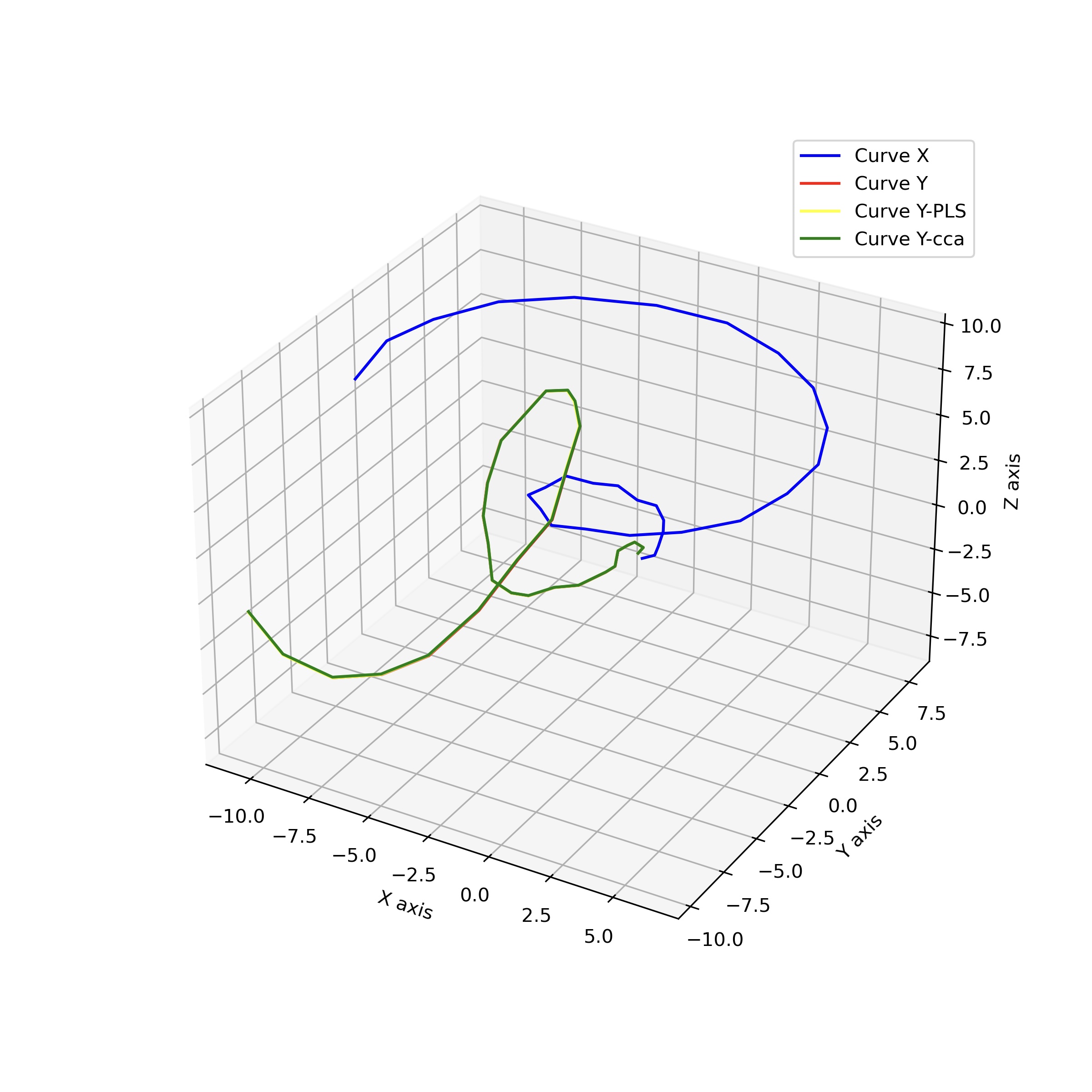}
        \caption{Without ground constraints}
        \label{fig4a}
    \end{subfigure}
    \hfill
    \begin{subfigure}{0.46\columnwidth}
        \centering
        \includegraphics[width=\linewidth]{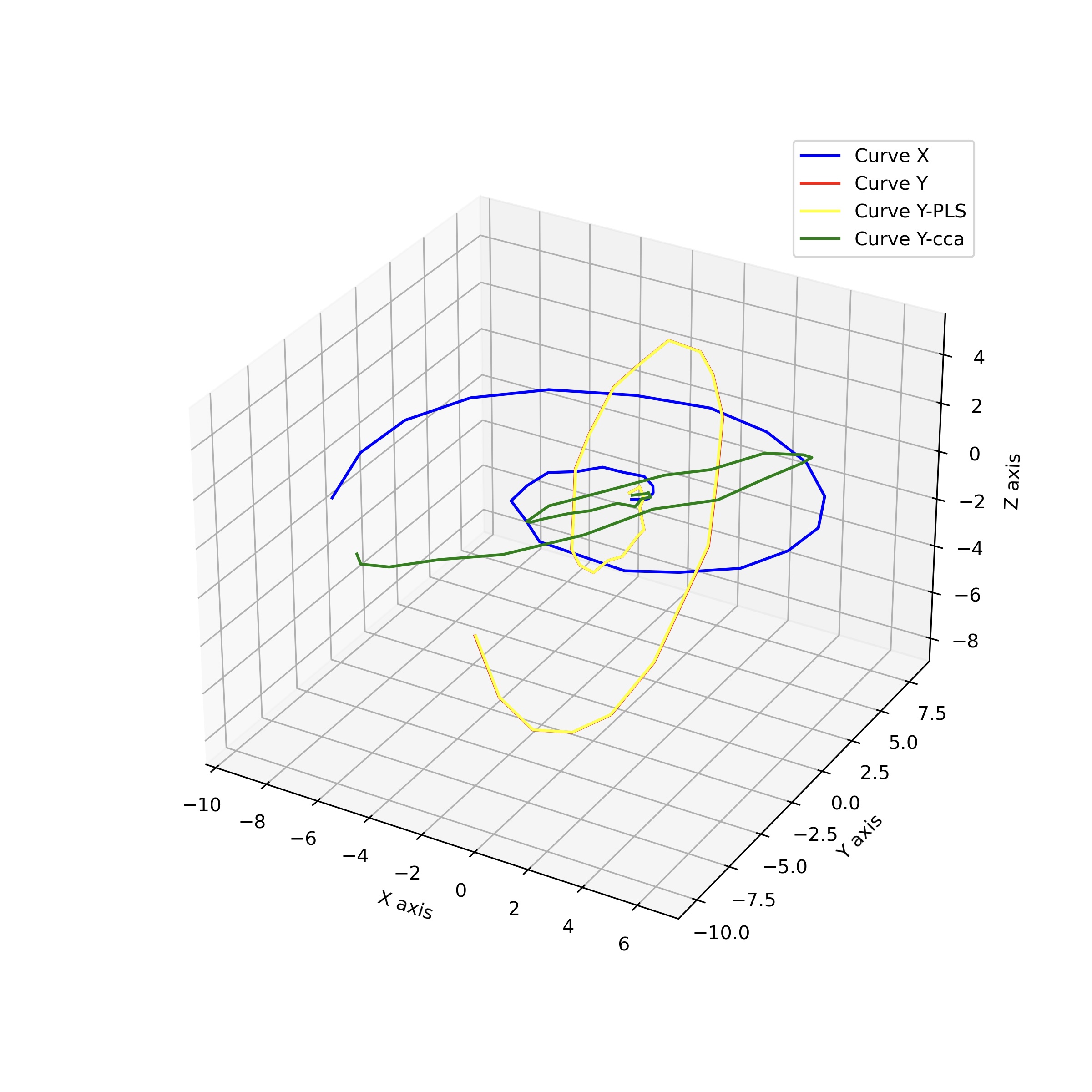}
        \caption{With ground constraints}
        \label{fig4b}
    \end{subfigure}

    \caption{Synthetic experiment results using PLS and CCA. (a) Full 6-DoF motion. (b) Ground-constrained planar motion.}
    \label{fig4}
\end{figure}

For each scenario, we created two sets of data representing the outputs of two sensors, with the second set generated using a known ground-truth rotation matrix $R_{gt}$. The goal is to recover the optimal rotation matrix $R$ between the two sensors. We applied both the PLS and CCA methods for calibration and compared the estimated rotation matrices $R_{pls}$ and $R_{cca}$ with $R_{gt}$ to assess the calibration accuracy. As is shown in Fig.~\ref{fig4}, we visualize the rotation relationship between the two sets of data by synthesizing the data, and the blue native data is curved from the real rotation matrix $R_{gt}$ to red. At the same time, the CCA method and the PLS method were used to solve the extrinsic rotation matrices for the two sets of data, respectively. 

In Fig.~\ref{fig4a}, the sensor can move in a full 6-DoF space, and the new trajectory estimated using the CCA method and the PLS method has a small error with the real trajectory, which almost coincides in the figure. However, when the ground constraints are received, the motion of the Z-axis cannot be obtained, and the CCA method has a large error in the calculation of the calculated and solved rotation matrix. In Fig.~\ref{fig4b}, the green trajectory obtained by $R_{cca}$ is offset by the yellow trajectory and the true red trajectory obtained by $R_{pls}$.

\begin{table}[ht]
  \centering
  \caption{Comparison of Error for Different Methods in Synthetic Data}
  \label{tab:error_comparison}
  \begin{tabular}{lcc}
    \toprule
    \textbf{Method} & \textbf{CCA~\cite{li2024spatio}} & \textbf{PLS(Ours)} \\
    \midrule
    Error(No Constraints) & \textbf{0.16} & \textit{0.21} \\
    \midrule
    Error(Ground Constraints) & \textit{26.56} & \textbf{0.26} \\
    \bottomrule
  \end{tabular}
\end{table}

\subsection{Robust Evaluation}

\begin{figure}[t]
    \centering

    \begin{subfigure}{0.4\columnwidth}
        \centering
        \includegraphics[width=\linewidth]{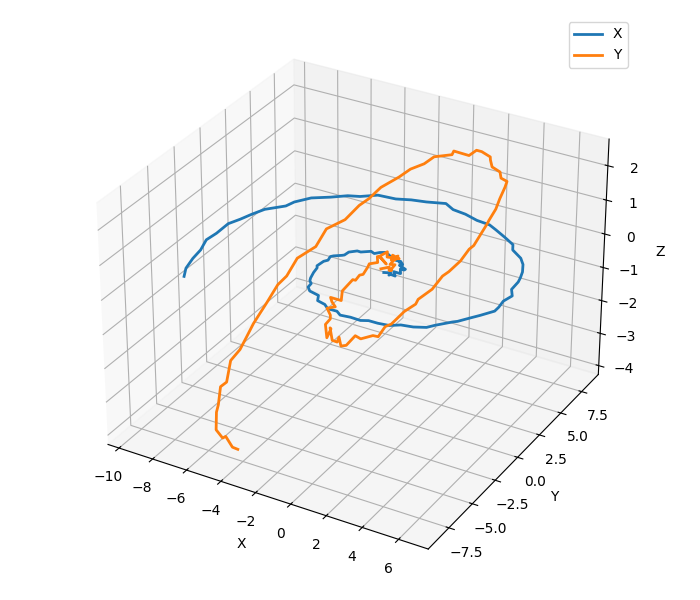}
        \caption{Circle}
        \label{fig:circle}
    \end{subfigure}
    \hfill
    \begin{subfigure}{0.4\columnwidth}
        \centering
        \includegraphics[width=\linewidth]{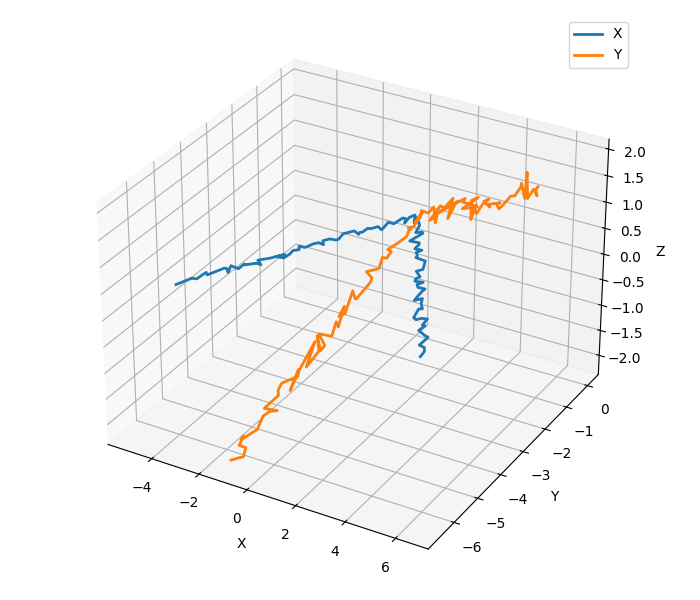}
        \caption{V-shape}
        \label{fig:v_shape}
    \end{subfigure}

    \vspace{3pt}

    \begin{subfigure}{0.4\columnwidth}
        \centering
        \includegraphics[width=\linewidth]{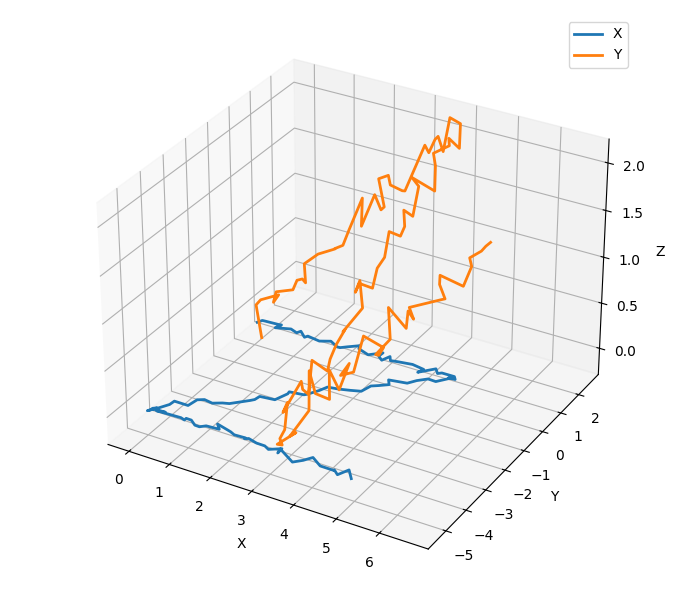}
        \caption{Z-shape}
        \label{fig:z_shape}
    \end{subfigure}
    \hfill
    \begin{subfigure}{0.4\columnwidth}
        \centering
        \includegraphics[width=\linewidth]{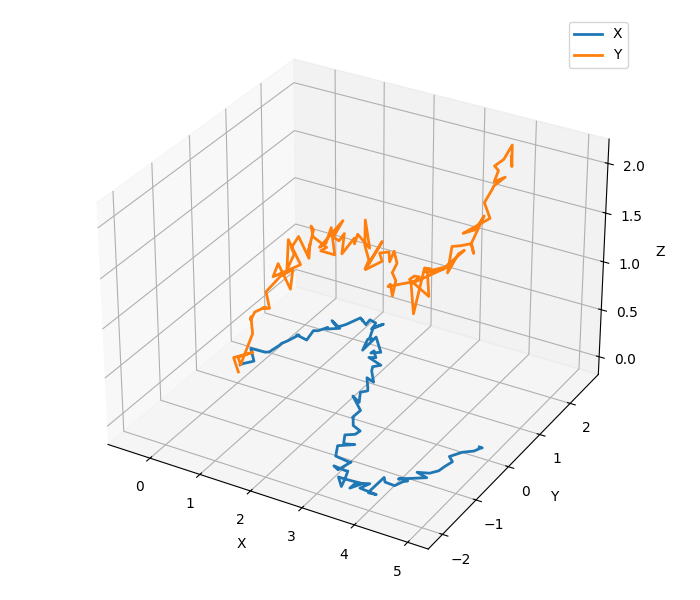}
        \caption{S-shape}
        \label{fig:s_shape}
    \end{subfigure}

    \caption{Illustration of the four planar motion trajectories used in the robustness evaluation.}
    \label{fig:routes}
\end{figure}

To evaluate the robustness of the proposed extrinsic rotation calibration method, we designed four different planar motion trajectories (Fig.~\ref{fig:routes}), namely \textit{Circle Route}, \textit{V-shape Route}, \textit{Z-shape Route}, and \textit{S-shape Route}. These trajectories were used to simulate paired 3D motion data generated by two sensors rigidly mounted on a robot moving along different planar patterns.



In each trial, a ground-truth rotation matrix $R_{\mathrm{gt}} \in SO(3)$ was randomly generated using ZYX Euler angles sampled uniformly within $[-90^\circ, 90^\circ]$. The first sensor's motion data $X$ were generated according to the predefined trajectory, with additive Gaussian noise to simulate measurement uncertainty. Consistently with the column-vector convention of Eq.~(6), the second sensor's data were constructed as
\begin{equation}
Y = X R_{\mathrm{gt}}^{\top} + \epsilon,
\end{equation}

where $\epsilon$ denotes zero-mean Gaussian noise with standard deviation $\sigma=0.10$, identical to that applied to $X$. Unless otherwise stated, all synthetic results in Table~II are reported at this noise level ($n=200$ trials, $100$ samples each).

Although the trajectories are spatially three-dimensional, the robot motion is predominantly constrained to planar movement, resulting in very limited excitation along the $z$-axis. Consequently, the rotation components associated with out-of-plane tilt (i.e., rotations around the $x$- and $y$-axes) become weakly observable. Under such conditions, the sample covariance matrix approaches rank deficiency in the degenerate direction, significantly increasing the condition number.

Methods that rely on covariance whitening, such as Canonical Correlation Analysis (CCA), are particularly sensitive to this degeneracy. When the covariance matrix becomes ill-conditioned, the whitening operation amplifies noise in poorly excited directions, leading to large rotation estimation errors. In contrast, the proposed method based on Partial Least Squares (PLS) does not require explicit whitening of degenerate covariance components. Instead, it directly maximizes cross-covariance alignment, thereby avoiding instability caused by near-singular covariance matrices.

As summarized in Table~\ref{tab:pls_cca_iros}, PLS consistently 
achieves high calibration accuracy across all motion patterns. 
Specifically, the mean rotation error of PLS remains below 
$1^\circ$ for the Circle, V-shape, and Z-shape trajectories, 
and approximately $1^\circ$ ($1.05^\circ$) for the S-shape 
trajectory. Compared with CCA, PLS reduces the estimation error 
by more than $94\%$ in all tested routes, with up to $99.5\%$ 
error reduction under the Circle trajectory.

These results confirm that the proposed method maintains high estimation accuracy even under low-excitation planar motion, while whitening-based approaches suffer from severe degradation due to covariance degeneracy.

\begin{table}[t]
\centering
\caption{Rotation Error Comparison Between PLS and CCA Across Different Routes (mean $\pm$ std, in degrees, n=200)}
\label{tab:pls_cca_iros}
\begin{tabular}{lccc}
\toprule
\textbf{Route} & \textbf{CCA~\cite{li2024spatio}} & \textbf{PLS (Ours)} & \textbf{Reduction (\%)} \\
\midrule

Circle   & 48.94 $\pm$ 0.10 & \textbf{0.23 $\pm$ 0.10} & 99.53 \\
V-shape  & 16.10 $\pm$ 0.20 & \textbf{0.48 $\pm$ 0.18} & 97.02 \\
Z-shape  & 49.95 $\pm$ 0.17 & \textbf{0.57 $\pm$ 0.23} & 98.86 \\
S-shape      & 20.33 $\pm$ 0.06 & \textbf{1.05 $\pm$ 0.47} & 94.83 \\
\bottomrule
\end{tabular}
\end{table}

\subsection{Real Data Collection}

We built a ground mobile robot platform to collect real-world sensor data. The system is based on a Jackal Clearpath robot equipped with odometry sensors and a mounted DAVIS 346 event camera. Data collection is handled by an onboard NVIDIA Orin module, and the calibration algorithm (implemented in Python) is executed on a PC with an Intel Xeon Silver 4410Y CPU.

Experiments were conducted in a flat environment. A circular calibration pattern was placed in front of the robot, which was manually controlled to approach the pattern from various angles. During motion, both event data and odometry readings were recorded. The collected data were then used to perform sensor calibration based on the trajectories from the two modalities.

Since motion capture is unavailable in our setup, we obtain a reference rotation using the intensity (frame-based) output of the event camera. Specifically, we apply a standard checkerboard-based calibration pipeline to the APS frames to estimate the extrinsic rotation between the camera and the robot. This frame-based calibration is performed independently from the proposed event-based method and serves as a reference for quantitative evaluation.
Since the APS frames and the event stream are produced by the same pixel array and share a common optical center~\cite{brandli2014davis}, the extrinsic rotation calibrated from the APS frames applies equally to the event stream, making it a geometrically valid reference. Although not obtained from an external tracking system, such frame-based calibration is widely adopted for event cameras and, being independent of the event-based pipeline under evaluation, provides a stable reference for assessing rotation accuracy.

\begin{figure}[h]
	
	\begin{minipage}{0.32\linewidth}
		\vspace{3pt}
		\centerline{\includegraphics[width=\textwidth]{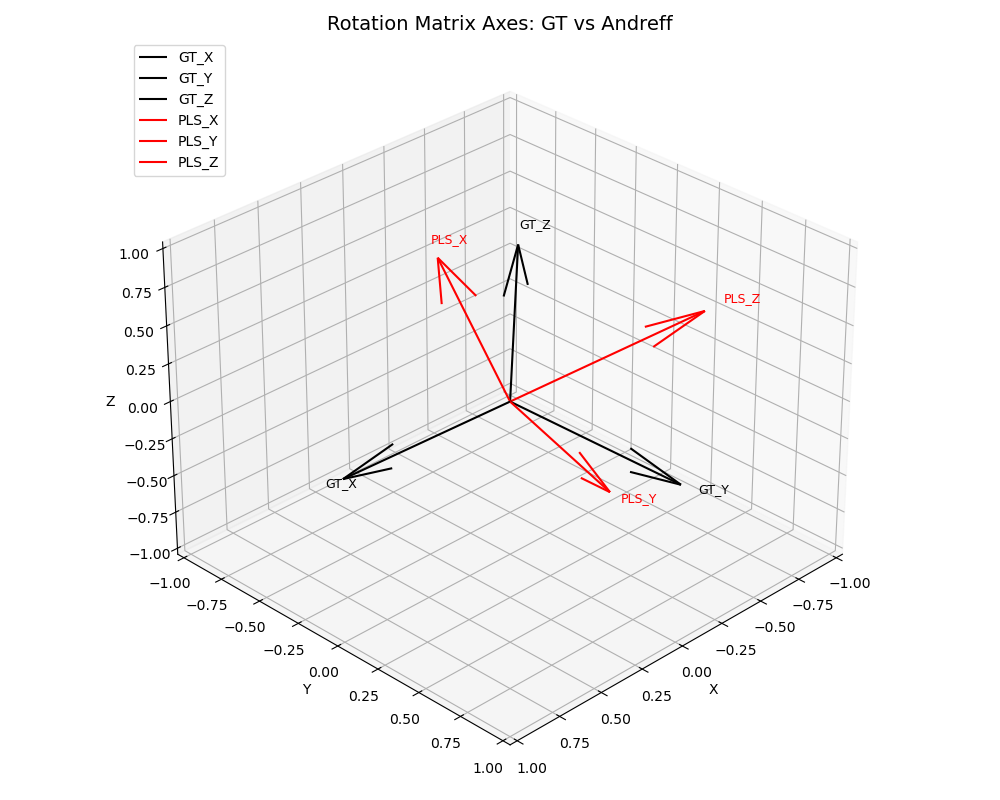}}
		\centerline{(a)}
	\end{minipage}
	\begin{minipage}{0.32\linewidth}
		\vspace{3pt}
		\centerline{\includegraphics[width=\textwidth]{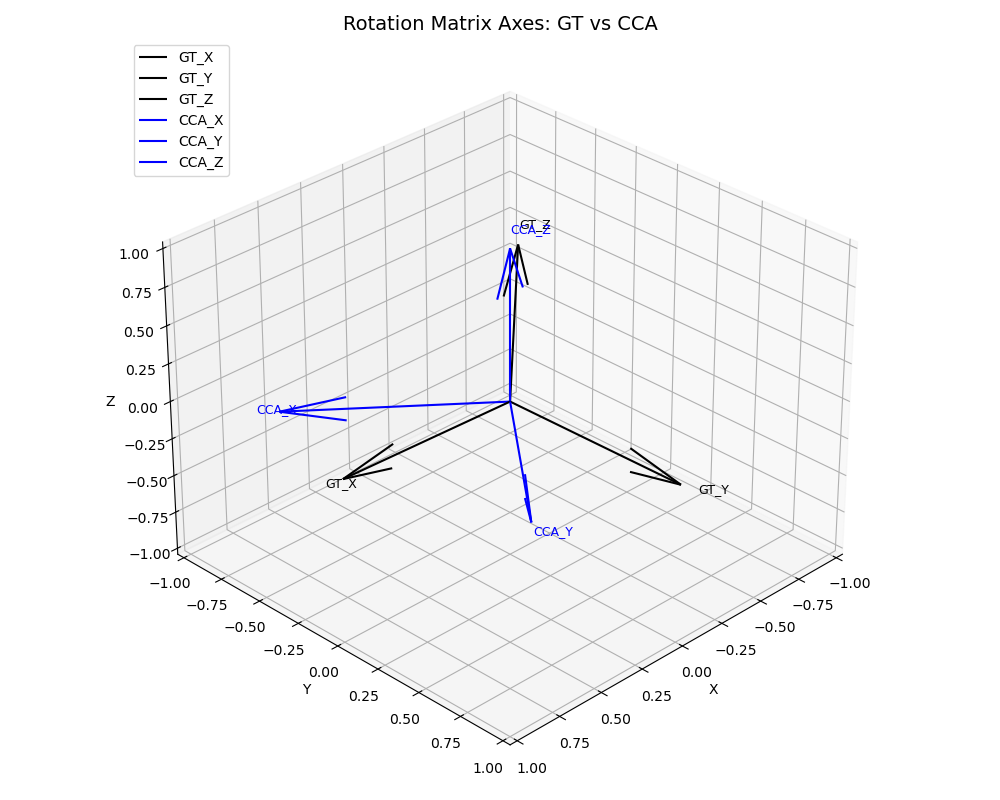}}
	 
		\centerline{(b)}
	\end{minipage}
	\begin{minipage}{0.32\linewidth}
		\vspace{3pt}
		\centerline{\includegraphics[width=\textwidth]{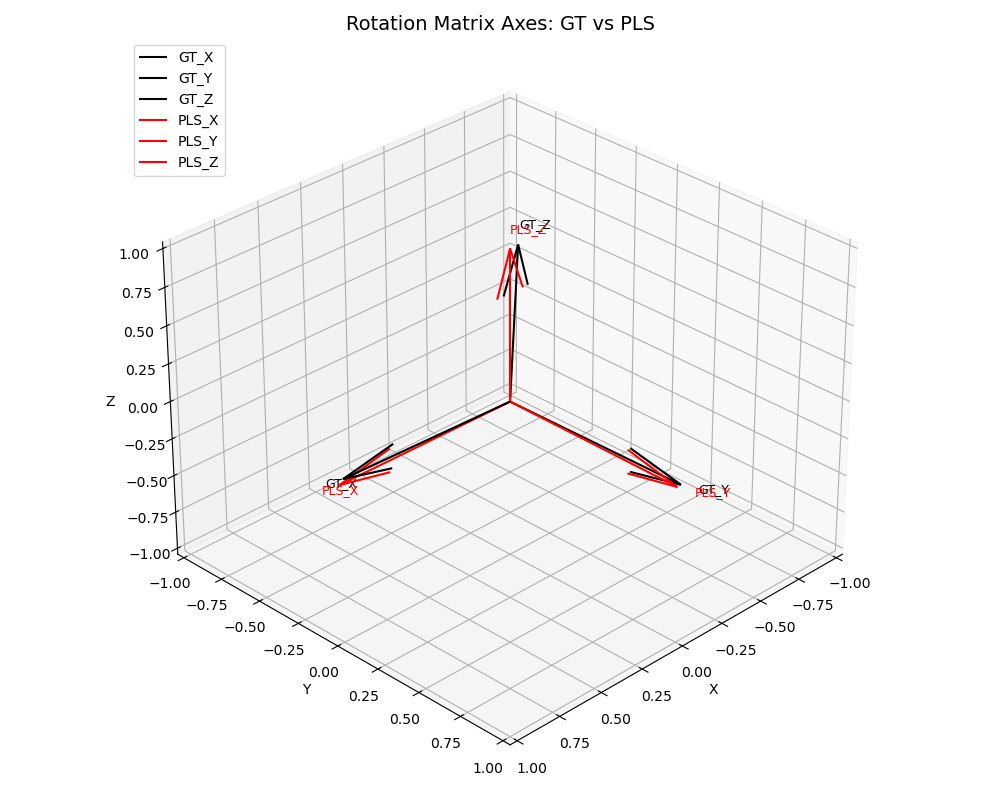}}
	 
		\centerline{(c)}
	\end{minipage}
 
	\caption{The visualization of calibration errors in three methods, (a) is Andreff, (b) is CCA-based and (c) is PLS-based method.}
	\label{fig6}
\end{figure}

In order to evaluate the performance of our method in the real world, we acquired event data and odometry data to compare the calibration errors in different methods, including Andreff solver\cite{wise2020certifiably}, CCA method and PLS method. We compare the external parameter estimation errors of three methods, as shown in Table~\ref{tab:real_error}. Both the Andreff and CCA methods exhibit varying degrees of deviation, while our proposed PLS method yields the smallest error. The estimated rotation matrices are visualized using Euler angles in Fig.~\ref{fig6}, where the PLS method produces results closest to the reference.

We measure rotation error by the geodesic distance on $\mathrm{SO}(3)$:
\begin{equation}
e_R = \arccos \left( \frac{\mathrm{Tr}\!\left(R_{\mathrm{est}} R_{\mathrm{gt}}^{\top}\right) - 1}{2} \right),
\end{equation}
reported in degrees.

\begin{table}[ht]
  \centering
  \caption{Comparison of Error for Different Methods in Real Data}
  \label{tab:real_error}
  \begin{tabular}{lccc}
    \toprule
     \textbf{Method} & \textbf{Andreff~\cite{wise2020certifiably}} & \textbf{CCA~\cite{li2024spatio}} & \textbf{PLS(Ours)} \\
    \midrule
    Error & \textit{74.65} & \textit{40.73} & \textbf{3.18} \\
    \bottomrule
  \end{tabular}
\end{table}

\section{CONCLUSIONS}

In this paper, we propose a novel event data representation and a more practical solution for sensor calibration in ground-constrained robotic systems. Conventional time-surface representations are heavily influenced by temporal decay dynamics, often leading to motion blur and reduced recognition accuracy in event-based vision. To address this, we introduce a polarity-based decay mechanism that mitigates temporal drag artifacts. This enhancement improves feature detection performance and provides a more reliable front-end for downstream sensor calibration. We also identify limitations in traditional CCA methods, where solving for 6-DoF motion under matrix singularities leads to inaccurate calibration results. To overcome this, we construct a PLS regression model that iteratively extracts latent components, enabling robust estimation of extrinsic rotation parameters while avoiding singularities. The proposed method is validated through experiments on both synthetic and real-world datasets. In the future, we hope that this method can provide some research inspiration for future sensor fusion systems based on event data and explore the possibilities of more topics.






\section*{ACKNOWLEDGMENT}

An LLM was used solely for language polishing assistance, and a generative AI tool was used only to create a schematic illustration of the robot platform. All scientific content and results were developed by the authors.



\Urlmuskip=0mu plus 1mu


\end{document}